\documentclass[conference,comsoc]{IEEEtran}
\usepackage{cite}

\ifCLASSINFOpdf
  \usepackage[pdftex]{graphicx}
\else
  \usepackage[dvips]{graphicx}
\fi
\usepackage[cmex10]{amsmath}
\usepackage{algorithmic}
\usepackage{array}
\ifCLASSOPTIONcompsoc
  \usepackage[caption=false,font=normalsize,labelfont=sf,textfont=sf]{subfig}
\else
  \usepackage[caption=false,font=footnotesize]{subfig}
\fi
\usepackage{dblfloatfix}
\usepackage{booktabs}
\usepackage{url}
\usepackage{float}
\usepackage{multirow}
\usepackage{enumerate}
\usepackage{amsthm}
\usepackage{threeparttable}
\usepackage{svg}
\usepackage{adjustbox}
\usepackage{booktabs,threeparttable}
\usepackage{array} % <-- new
\newcolumntype{C}[1]{>{\centering\arraybackslash}p{#1}} % <-- new

\newtheoremstyle{mystyle}%                % Name
  {}%                                     % Space above
  {}%                                     % Space below
  {\itshape}%                                     % Body font
  {}%                                     % Indent amount
  {\bfseries}%                            % Theorem head font
  {.}%                                    % Punctuation after theorem head
  { }%                                    % Space after theorem head, ' ', or \newline
  {}%                                     % Theorem head spec (can be left empty, meaning `normal')
\theoremstyle{mystyle}

\newenvironment{talign*}
 {\csname align*\endcsname}
 {\endalign}
\usepackage{arydshln}
\usepackage[normalem]{ulem}
\usepackage{hyperref}
\usepackage{amsmath}
\usepackage{xurl}

\begin{document}

%\title{Explainable AI (XAI) for Detection of Adversarial Attacks on Neural Networks\\
\title{Explainable Machine Learning for Cyberattack Identification from Traffic Flows}

\author{Yujing Zhou$^{1a}$, Marc L. Jacquet$^{1a}$, Robel Dawit$^{1a}$, Skyler Fabre$^{1a}$, Dev Sarawat$^{1a}$,\\ Faheem Khan$^{1a}$, Madison Newell$^{1a}$, Yongxin Liu$^{1a}$, Dahai Liu$^{1a}$, Hongyun Chen$^{1a}$, Jian Wang$^{2b}$,, Huihui Wang$^{3c}$\\
    % Wenkai Tan$^{~1a}$, Justus Renkhoff$^{~2b}$, Alvaro Velasquez$^{3c}$, Jian Wang$^{4e}$\\ Shuteng Niu$^{5f}$, Yongxin Liu $^{1d}$, Houbing Song$^{2b}$\\
$^{1}$Embry-Riddle Aeronautical University, FL 32114 USA,\\
$^{2}$University of Tennessee at Martin, TN 38238 USA, $^{3}$Northeastern University, Arlington, VA 22209\\
    $^{a}$\{zhouy9, jacquetm,  dawitr, fabres, sarawatd, khanf10, newellm2\}@my.erau.edu, \\$^{a}$liuy11,liu89b,chenh4@erau.edu
    $^{b}$jwang186@utm.edu, $^{c}$huih.wang@northeastern.edu\\
}
\IEEEtitleabstractindextext{%
\begin{abstract}

The increasing automation of traffic management systems has made them prime targets for cyberattacks, disrupting urban mobility and public safety. Traditional network-layer defenses are often inaccessible to transportation agencies, necessitating a machine learning-based approach that relies solely on traffic flow data. In this study, we simulate cyberattacks in a semi-realistic environment, using a virtualized traffic network to analyze disruption patterns. We develop a deep learning-based anomaly detection system, demonstrating that Longest Stop Duration and Total Jam Distance are key indicators of compromised signals. To enhance interpretability, we apply Explainable AI (XAI) techniques, identifying critical decision factors and diagnosing misclassification errors. Our analysis reveals two primary challenges: transitional data inconsistencies, where mislabeled recovery-phase traffic misleads the model, and model limitations, where stealth attacks in low-traffic conditions evade detection. This work enhances AI-driven traffic security, improving both detection accuracy and trustworthiness in smart transportation systems. All implementation details and source code are publicly available on GitHub at: https://github.com/U1overground/Cybersummer

\end{abstract}

}

\IEEEoverridecommandlockouts
\maketitle
\IEEEdisplaynontitleabstractindextext
\IEEEpeerreviewmaketitle

\section{Introduction}

The increasing automation and connectivity of transportation systems have heightened the risk of cyberattacks, posing threats to traffic flow, public safety, and operational continuity \cite{vegesna2022investigations}. Among these systems, traffic management infrastructure—particularly traffic lights—has become a prime target due to its critical role in urban mobility. Cyberattacks on traffic control can disrupt emergency responses, delay public transit, and cause significant economic losses \cite{feng2022cybersecurity, ozarpa2021cyber}. As cities adopt smart traffic control solutions, securing these systems against cyber threats is of great significance.

Existing approaches to transportation cybersecurity often focus on network-layer defenses or direct intrusion detection using Machine or Deep Learning (ML/DL) models. These methods involve analyzing network traffic for irregular patterns indicative of potential cyberattacks, with algorithms like support vector machines, random forests, and deep learning models being implemented to enhance detection and response capabilities. In addition, although Deep Learning methods have shown great potential in cyberattack detection but its blackbox nature makes it hard to be applied in safety-critical scenarios like traffic control systems. To counter this issue, Explainable AI (XAI) technologies provide insights into how machine learning models make predictions, enhancing interpretability and trust. Common XAI techniques include feature attribution methods, such as SHAP (Shapley Additive Explanations) and LIME (Local Interpretable Model-agnostic Explanations), which quantify the contribution of individual features to a model’s decision. Occlusion sensitivity and Grad-CAM are widely used for deep learning models, highlighting the most influential input regions by systematically masking or tracking gradient responses.

However, data packets from the network layer are not always available to transport management agencies, as traffic control facilities may rely on public cellular networks for data communication. This challenge on one hand, demands an alternative detection approach that uses machine learning on traffic data alone to identify compromised traffic signaling devices. On the other hand, explanability are crucial to maintain the trustworthiness of machine learning models.

This research simulates the traffic network in a semi-realistic city environment, capturing the responses traffic flows when traffic lights are being compromised by the adversaries. In this scenario, each traffic light is simulated as a virtual machine connected via an internal network, allowing realistic interactions and cyberattack scenarios. We apply deep neural network (DNN) models to detect anomalies in traffic patterns and used XAI to provide a thorough analysis on the models' behaviors, offering a secondary defense layer against subtle cyberattacks on traffic signals. Our contributions are as follows:
\begin{itemize}
    \item Developed a deep learning-based anomaly detection system to identify compromised traffic signals using traffic data alone.
    \item XAI methods reveal that \textit{Longest Stop Duration} and \textit{Total Jam Distance} are key indicators of cyberattacks.
    \item Misclassification analysis identifies two error sources: transitional data issues and model limitations in stealth attacks under low-traffic conditions.
\end{itemize}

\section{Related Work}
\label{sectRW}

Transportation systems are increasingly targeted in cybersecurity attacks due to their critical role in public infrastructure and safety. Research highlights that interconnected traffic control systems are particularly vulnerable due to their networked architecture, remote access points, and integration with external devices and sensors—factors that adversaries can exploit to manipulate traffic patterns and disrupt operations \cite{pundir2022cyber}. The shift toward automated traffic management introduces additional security challenges, especially in dense urban areas where the complexity of traffic flow amplifies the potential impact of cyberattacks \cite{feng2018vulnerability, zhou2022survey}. Given the financial and safety risks posed by such disruptions, researchers stress the need for robust cyber-defense frameworks specifically designed for traffic management infrastructure \cite{ma2021smart}.

Cyberattacks on transportation Cyber-Physical Systems (CPS) can be broadly classified into several categories. Denial of Service (DoS) attacks flood traffic control networks with excessive requests, disrupting operations and leading to congestion and delays \cite{gu2007denial}. Data manipulation attacks involve unauthorized modifications to sensor data, resulting in incorrect signal timings and misinformed traffic decisions \cite{chen2016analysis}. Replay attacks allow adversaries to resend previously intercepted valid data, causing outdated or misleading instructions to unpredictably affect traffic flow \cite{syverson1994taxonomy}. Command injection attacks enable attackers to introduce malicious commands into traffic control systems, potentially manipulating signals or creating hazardous scenarios, such as synchronizing all lights to green or red at intersections \cite{usama2024command}. Man-in-the-middle (MitM) attacks intercept and alter real-time communication between sensors and controllers, enabling attackers to control or distort critical operational data \cite{conti2016survey}. These attack vectors exploit vulnerabilities in both cyber and physical layers, highlighting the urgent need for integrated security strategies in transportation CPS.

Detecting cyberattacks in transportation Cyber-Physical Systems (CPS) requires methods capable of identifying anomalies across both cyber and physical layers, as these systems integrate digital signals with real-world consequences. Common techniques include machine learning-based anomaly detection and supervised classification, which analyze network traffic, sensor data, and system commands to recognize unusual patterns indicative of attacks \cite{bhuyan2013network}. Recent studies highlight the effectiveness of Random Forest, neural networks, and ensemble models in detecting complex anomalies, such as manipulated command sequences and abnormal network activity spikes \cite{corea2024explainable}. Given the real-time nature of CPS, many detection systems incorporate real-time analytics to issue immediate alerts upon detecting deviations \cite{liu2021zero}. Additionally, hybrid approaches combining rule-based detection with deep learning have proven effective in managing high-dimensional data and addressing the challenge of imbalanced CPS traffic, where attack events are rare but critical \cite{liu2021class,pang2021deep}.

Beyond conventional cybersecurity methods for Cyber-Physical Systems, traffic flow analysis is an emerging approach for indirectly detecting cyberattacks in traffic control systems. This method identifies anomalies in traffic behavior—such as signal tampering or coordinated disruptions—that may indicate malicious activity. Research has shown that unusual spikes in metrics like vehicle occupancy and queue length can signal compromised traffic signals or network intrusions \cite{jyothi2024data}. However, the effectiveness of this approach depends on the complexity of traffic patterns and the subtlety of the attack, as more sophisticated disruptions may blend seamlessly into normal fluctuations \cite{bawaneh2019anomaly}.

\section{Methodology}
\label{sectMM}
\subsection{Simulation Framework}
We simulate cyberattacks and system responses within the traffic control system of a semi-realistic city. Our dataset includes base map tiles, traffic light locations, and cellular tower positions, assuming that traffic light controllers rely on these for internet connectivity. In addition, we incorporate annual average daily traffic volumes for major roads to improve modeling accuracy. To make the simulation as realistic as possible, the following key technologies are employed:

    \begin{itemize}
        \item \textit{Infrastructure Layer:} Raspberry PI Virtual Machines are used to seamlessly simulate the hardware and software environment of traffic control devices. We believe that the code in such scenario is practically deployable.
        \item \textit{Traffic Dynamic Simulation Layer}: Each Raspberry PI VM simulates one traffic light controller and connects remotely to \textit{SUMO} (https://eclipse.dev/sumo/) simulator via the \textit{TraCI} (https://sumo.dlr.de/docs/TraCI.html) interface. It is assumed that each traffic light controller only controls one intersection. We assume that each traffic light controller updates its phase pattern every 10 seconds.        
    \end{itemize}
With the help of this simulation framework, we can directly use the \textit{Metasploit} (https://www.metasploit.com/) framework and OpenVAS (https://www.openvas.org/) to simulate cyber-attack and defense events.

\subsection{Attack Scenarios and Data Collection}

\begin{figure}[h]
    \centering
    \includegraphics[width=0.9\linewidth]{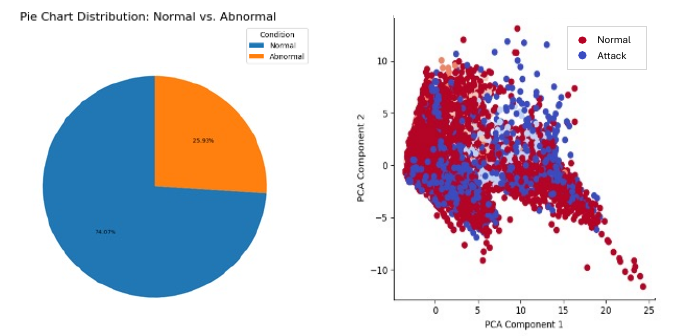}
    \caption{Data distribution of the simulated attack and normal scenarios. We used Principal Component Analysis (PCA) to project the original data to 2D space with 90\% of variance explained.}
    \label{figSimulationDataDistrib}
\end{figure}

We simulated the attack scenario that traffic lights are turned all green or all red randomly in the busiest intersection of the target city for one hour. Various traffic flow statistics are collected every 10 seconds, and these collected statistics form the initial dataset for an attack scenario. The distribution of the data w.r.t. different categories are shown in Figure~\ref{figSimulationDataDistrib}. 

As depicted, inferring cyber-attacks from traffic patterns could be challenging owing to two main factors: a) The dataset if highly imbalanced as the chance of observing a cyber-attack is rare and the collected dataset are highly imbalanced even after Synthetic Minority Over-sampling Technique (SMOTE) algorithm is applied to re-balance the dataset. b) As in Figure~\ref{figSimulationDataDistrib}, the data tuples representing attack scenarios are mixed almost uniformly with the normal scenarios in the 2D space generated by PCA.

% The data were generated to simulate four distinct types of hack, together with a control simulation representing scenarios where no hack occurred. \textcolor{blue}{Which four types of attacks?}} 

For simplicity, all categories of hacked data were consolidated into a single class. This allowed for the training of binary classifiers capable of distinguishing between hacked and non-hacked states. To prevent learning trivial features, the numeric data generated from SUMO were normalized, meanwhile, irrelevant features, including target, begin, end, and id, were removed. 

\subsection{Convolutional Neural Network}
We implemented a convolutional neural network (CNN) to further improve the classification of traffic data and detect potential hacked instances. By exploring three different input matrix configurations: 9x23 (5 seconds), 18x23 (10 seconds), and 36x23 (20 seconds). And we aimed to identify the most suitable temporal resolution for detecting abnormalities from these datasets.

\subsubsection{Data Preparation}

The input features were normalized using MinMaxScaler, scaling their values between 0 and 1. Depending on the selected time interval, the data was structured into matrices of three sizes: 9×23, 18×23, and 36×23. Each matrix represents a different time window (5, 10, or 20 seconds), where rows correspond to sensor data, and columns capture 23 distinct traffic-related features. Using different length of time window, we can explore whether shorter or longer observation windows affect the network’s ability to detect hacking activities in the traffic data. 

In addition to the single-layer matrix inputs, we explored a three-layer tensor configuration that incorporated the original traffic matrix alongside mean and standard deviation matrices, computed over 10-second intervals from the control dataset. This multi-layer tensor aimed to provide additional statistical features that potentially enhancing the CNN’s ability to detect deviations from normal traffic patterns.

The target labels were converted into a binary format, with 1 indicating normal traffic (control group) and 0 representing hacked traffic. Labels were extracted at regular intervals matching the matrix size to maintain alignment between inputs and their corresponding labels. The processed datasets were then split into training and testing sets using an 80-20 split and converted into PyTorch tensors for CNN training.

\subsubsection{Network Architecture}
\begin{figure}[h]
    \centering
    \includegraphics[width=\linewidth]{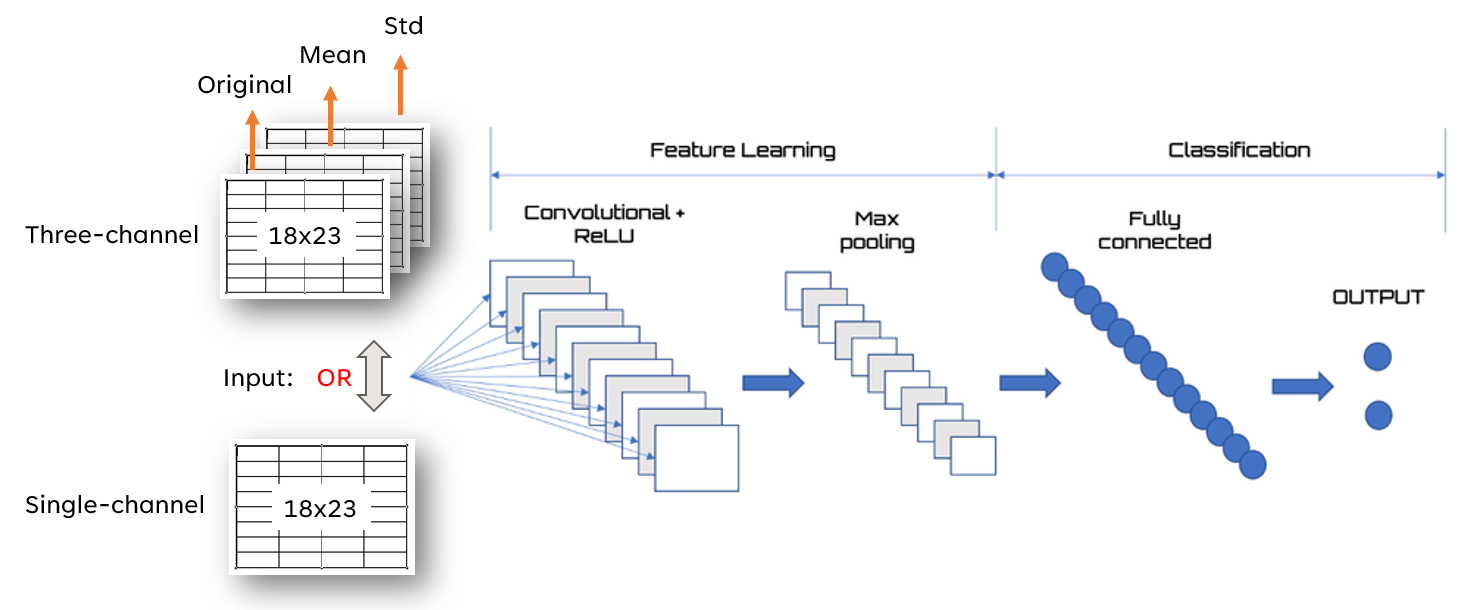}
    \caption{Architecture of the Convolutional Neural Network (CNN)}
    \label{FigCNNArch}
\end{figure}

Our network architecture is illustrated in Figure~\ref{FigCNNArch}. The input layer is adapted to accommodate varying time window lengths, ensuring compatibility across different input sizes. The core architecture consists of two convolutional layers with 64 and 128 filters, respectively, each employing a 3×3 kernel. Following each convolutional layer, a max-pooling layer with a 2×2 kernel reduces spatial dimensions while retaining critical features. The pooled outputs are then flattened, with dimensions dynamically computed to ensure seamless integration of all three possible input sizes into the fully connected layers. The fully connected section includes a hidden layer with 64 neurons, followed by a single output neuron utilizing a sigmoid activation function for binary classification.

The model was trained using Binary Cross-Entropy Loss to minimize the discrepancy between predicted probabilities and true labels. The Adam optimizer was used with a learning rate of 0.001, and training was conducted for 10 epochs using mini-batches of size 32. To enhance generalization, data was shuffled at the start of each epoch. The model was evaluated in three different input dimensions (9×23, 18×23, and 36×23) to analyze the impact of varying observation windows on performance. 

% \begin{table}
% \centering
% \caption{Confusion Matrices Analysis}
% \label{tab:confusion_matrices}
% \resizebox{\linewidth}{!}{%
% \begin{tabular}{lrrrrrr}
% \toprule
% Scenario  &    TN &    FP &   FN &    TP &  Accuracy &  F1-Score \\
% \midrule
% 5-second  &  1233 &  1679 &  612 &  5718 &     75.21 &      0.83 \\
% 10-second &   932 &   485 &  371 &  2833 &     81.48 &      0.87 \\
% 20-second &   246 &   443 &   56 &  1566 &     78.41 &      0.86 \\
% \bottomrule
% \end{tabular}%
% }
% \end{table}

% \begin{figure*}[!t]
%     \centering
%     \includegraphics[width=2\columnwidth]{CNNconfusion.png}
%     \caption{Confusion Matrix of CNN with Various Input Sizes}
%     \label{fig:CNNconfusion}
% \end{figure*}

\subsection{Explainable AI}
We applied two widely used explainable AI (XAI) techniques, occlusion sensitivity \cite{zeiler2014visualizing} and local interpretable model-agnostic explanations (LIME) \cite{ribeiro2016why} to gain deeper insight into the models' decision-making processes. These techniques help identify key features influencing predictions, enhancing interpretability and transparency in complex model behavior.

\subsubsection{Occlusion Sensitivity Analysis}
Occlusion Sensitivity systematically removed individual features from the input to measure their impact on model performance, which is achieved by systematically masking regions of the input and observing how the model's output changes \cite{zeiler2014visualizing}. The sensitivity score is computed as:
\[
\Delta P = \left| P(y=1 \mid x) - P(y=1 \mid x_{\text{occluded}}) \right|
\]
where \(P(y=1 \mid x)\) is the original prediction probability and \(P(y=1 \mid x_{\text{occluded}})\) is the probability after masking part of the input. The computed feature importance scores revealed which variables had the greatest impact on predictions, establishing a baseline for comparison with the CNN's spatial sensitivity.

\subsubsection{LIME:} In our study, LIME highlighted how each feature influenced predictions for specific cases. It learns a simpler local model, like a linear model or a small decision tree, to approximate how the original model behaves around the data point being explained. By creating slightly modified versions of that data point and observing changes in the black box model’s output, LIME identifies which features have the most impact in that local area \cite{ribeiro2016why}. To use LIME, we flattened the input tensors into one-dimensional arrays to meet its requirements. Although this step removed spatial context, LIME still pinpointed the features most critical to the CNN’s predictions. 

\section{Results}
\label{sectEED}

\subsection{Model Performance and Feature Explanation}

As in Table~\ref{tab:cnn_performance}, using a 5-second window and one-layer matrix leads to a relatively high number of incorrect classifications, particularly where abnormal signals are labeled as normal. This suggests that such a short time frame may not capture enough information for reliably identifying hacked conditions. Increasing the window to 10 seconds (18×23) substantially reduces both false negatives and false positives, resulting in an improved accuracy of 81.48\%. Extending the window further to 20 seconds (36×23) moderates some misclassifications but lowers the accuracy slightly to 78.41\%, likely reflecting that longer time spans may introduce unnecessary complexity. Therefore, a 10s window appears to keep the balance and provide sufficient detail to detect sudden anomalies without overwhelming the model.
% \begin{table}[h]
% \centering
% \caption{Performance of CNN model with various inputs}
% \label{tab:confusion_matrices}
% \resizebox{\linewidth}{!}{%
% \begin{tabular}{lrrrrrr}
% \toprule
% Scenario  &  TNR &  FPR &  FNR &  TPR &  Acc.\% &  F1 \\
% \midrule
% 5-second  &  0.42 &  0.58 &  0.10 &  0.90 &  75.21 &  0.83 \\
% 10-second &  0.66 &  0.34 &  0.12 &  0.88 &  81.48 &  0.87 \\
% 20-second &  0.36 &  0.64 &  0.03 &  0.97 &  78.41 &  0.86 \\
% \bottomrule
% \end{tabular}%
% }
% \end{table}

\begin{table}[h]
\centering
\caption{Performance of the CNN Model under Different Input Configurations}
\label{tab:cnn_performance}
\resizebox{\linewidth}{!}{%
\begin{tabular}{lcccc}
\toprule
\textbf{Configuration} & \textbf{Accuracy (\%)} & \textbf{Precision} & \textbf{Recall} & \textbf{F1-Score} \\
\midrule
5-second      & 75.21 & 0.74 & 0.75 & 0.73 \\
\textbf{10-second}     
              & \textbf{81.48} & \textbf{0.84} & \textbf{0.92} & \textbf{0.88} \\
20-second     & 78.00 & 0.79 & 0.78 & 0.75 \\
Three-layer   & 73.00 & 0.71 & 0.73 & 0.71 \\
\bottomrule
\end{tabular}
}
\end{table}

\subsubsection{Occlusion Sensitivity}

\begin{figure}
    \centering
    \includegraphics[width=0.9\linewidth]{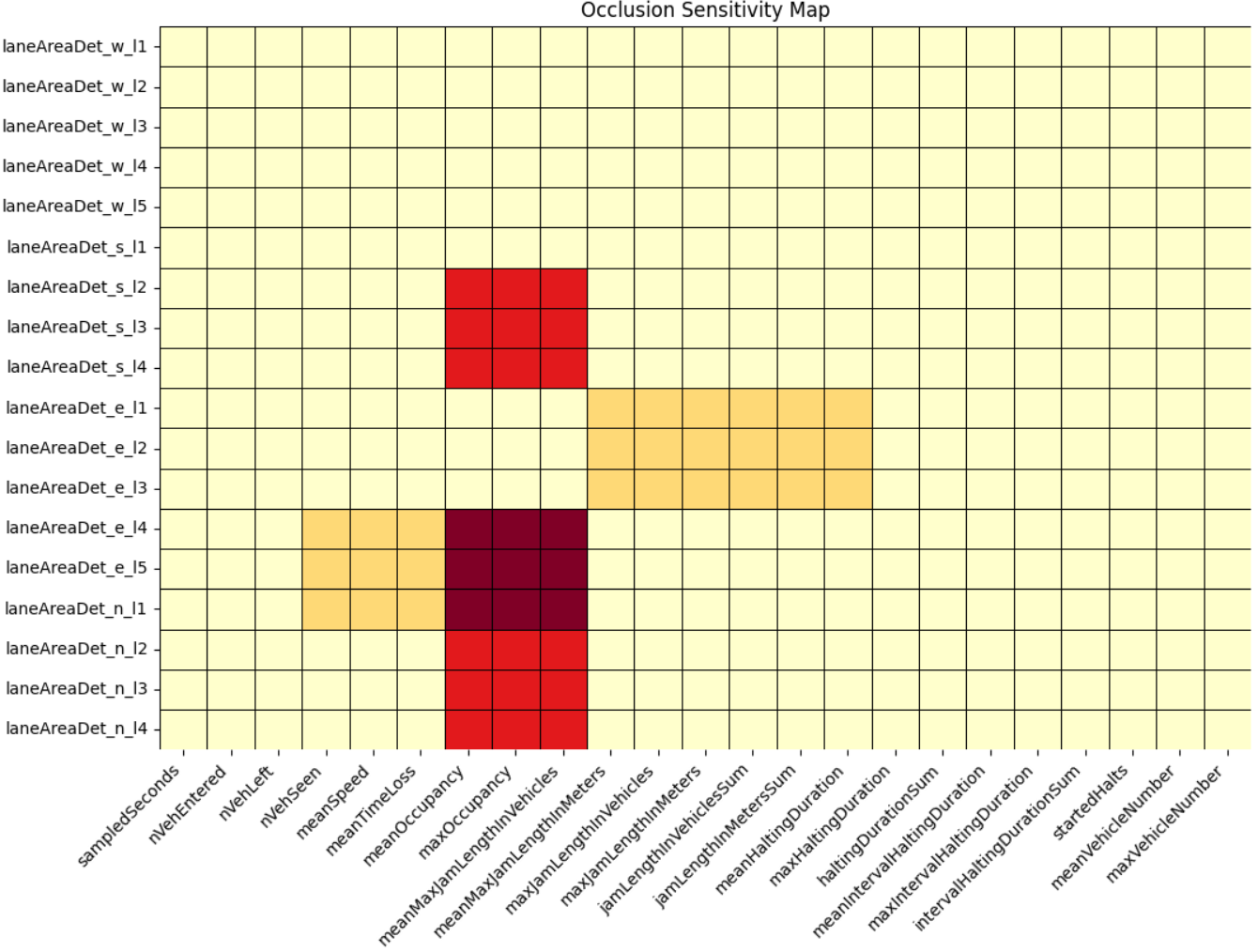}
    \caption{Occlusion Sensitivity of CNN}
    \label{CNN_OCC}
\end{figure}

As in Figure~\ref{CNN_OCC}, each cell represents the impact of occluding a specific combination of sensor features within our 10-second CNN model. Warmer tones (red) indicate a greater drop in accuracy, suggesting that the network relies heavily on those inputs for classification. Lighter colors (yellow) show regions where occlusion has only a minor effect, indicating less influence on the final prediction. The map indicates that CNN is particularly sensitive to certain lane detectors (e.g., those facing south or north) and features capturing congestion severity, implying that the model prioritizes signals of unusual traffic build-up when identifying hacked conditions.

\begin{figure}[htb]
    \centering
    
    % First subfigure (Normal Case)
    \subfloat[]{
        \includegraphics[width=0.95\linewidth]{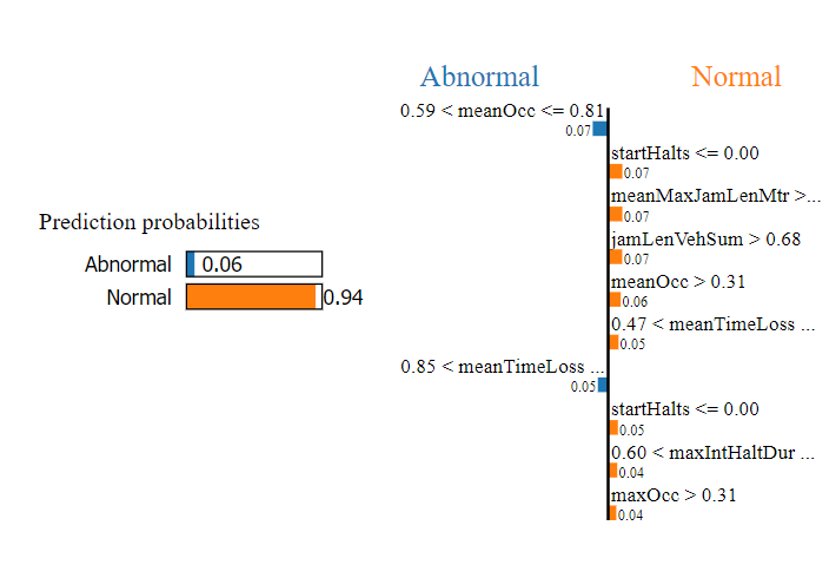}
        \label{fig:lime_cnn_normal}
    }
        
    % Second subfigure (Hacked Case)
    \subfloat[]{
        \includegraphics[width=0.95\linewidth]{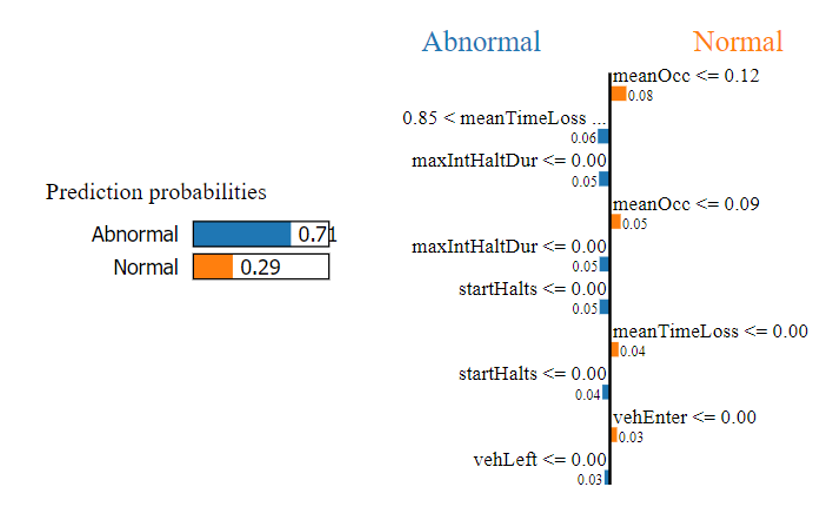}
        \label{fig:lime_cnn_hacked}
    }
    
    \caption{LIME Explanations for Two Different Traffic Scenarios:
    (a) Normal Traffic and (b) Hacked Case.}
    \label{fig:lime_cnn_compare}
\end{figure}

\subsubsection{LIME}
LIME further explain how the CNN arrives at its predictions. As exemplified in Figure\ref{fig:lime_cnn_compare}(a), the CNN assigns a high confidence of approximately 92\% to the normal label. The LIME bar chart indicates that moderate occupancy values (e.g., meanOcc around 0.3) and relatively short jam lengths (e.g., jamLenMtr below 50m) provide positive weights of about +0.25 and +0.20, respectively, contributing strongly toward a normal classification. In contrast, features such as higher maximum occupancy or jam length would have make the model predict in the opposite direction.

Moreover, the explanation for a hacked traffic sample (Figure\ref{fig:lime_cnn_compare}(b)) emphasizes the opposite pattern: unusually high congestion metrics, such as elevated jam lengths and prolonged halting durations, dominate the feature contributions. These features push the CNN to classify the scenario as “abnormal” or "hacked." Notably, certain threshold values for key features (e.g., mean halting duration exceeding a specific limit) appear consistently in these explanations, reinforcing the conclusion that severe or prolonged traffic buildup is a primary indicator of hacking activity.

\subsection{Misclassifications Analysis}

Our best performing CNN model achieved 81\% accuracy, but still misclassifies hacked signals in approximately 19\% of the test cases. To identify the underlying causes of these errors, we applied SHAP Analysis to the misclassified samples. Our objective was to determine whether the errors comes from data-related issues, such as typical traffic patterns incorrectly labeled as normal, or from model limitations, where CNN fails to recognize specific types of hack.

\subsubsection{Transitional Data Issues}
\begin{figure}[h]
    \centering
    \subfloat[]{%
        \includegraphics[width=0.9\columnwidth]{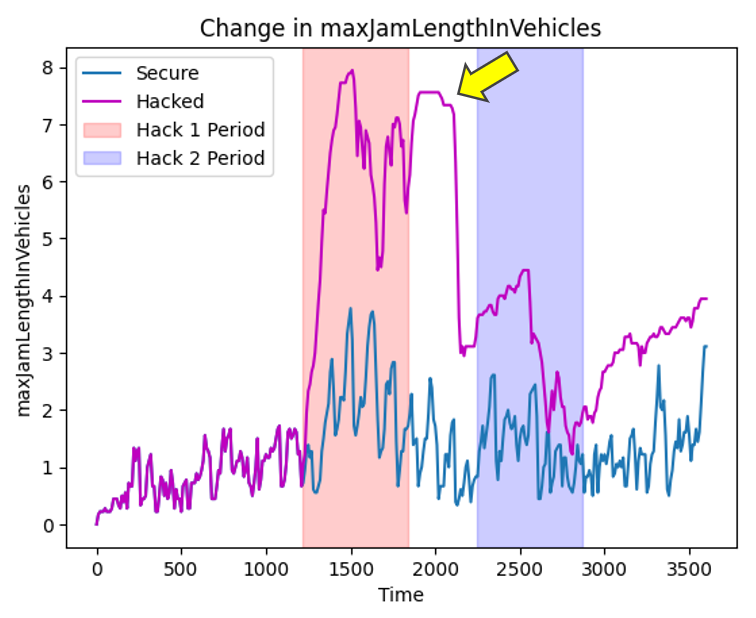}%
        \label{fig_data1}
    }
    \\
    \subfloat[]{%
        \includegraphics[width=0.9\columnwidth]{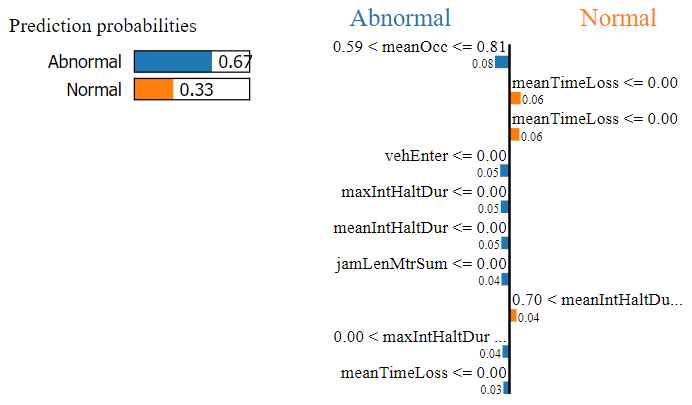}%
        \label{fig_data2}
    }
    \caption{Transitional Data Issues: (a) data remain abnormal while hack was ended. (b) SHAP analysis}
    \label{fig_data}
\end{figure}

One source of error occurs when traffic has not fully stabilized following a major event or partial road closure. Although the data is labeled as normal, its patterns appear atypical, leading the model to detect congestion or halting behaviors that deviate from typical normal traffic, as shown in Figure~\ref{fig_data1}. Explainable AI (XAI) analysis reveals that the CNN assigns greater importance to moderate occupancy signals, as they generally indicate normal conditions (Figure~\ref{fig_data2}). However, in reality, the traffic is still in a recovery phase, meaning the normal label may not fully reflect actual conditions. Thus, this type of error comes from labeling inaccuracies: the system is not yet truly back to normal, causing the model to misinterpret transitional traffic states.

\subsubsection{Model Errors}
\begin{figure}[!ht]
    \centering
    \subfloat[]{%
        \includegraphics[width=0.9\columnwidth]{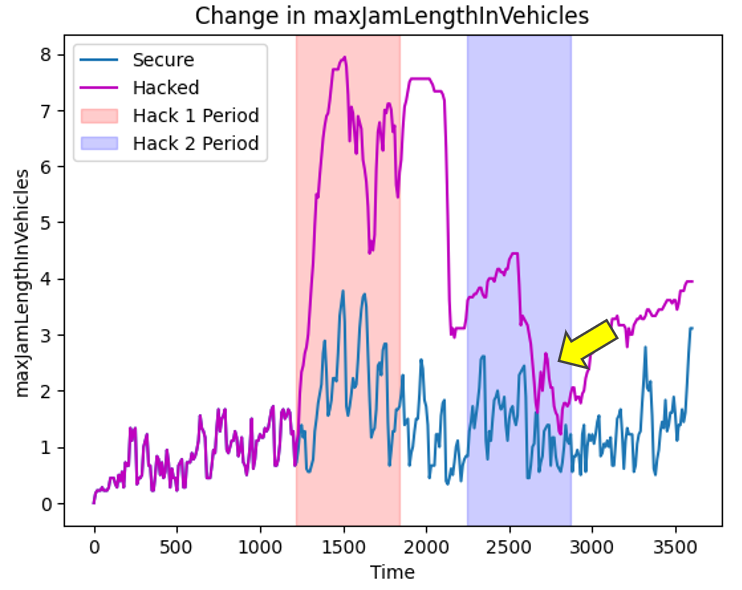}%
        \label{fig_model1}
    }
    \\
    \subfloat[]{%
        \includegraphics[width=0.9\columnwidth]{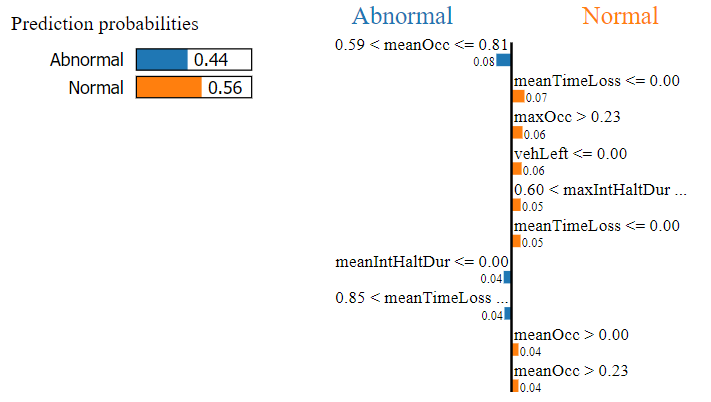}%
        \label{fig_model2}
    }
    \caption{Model Errors: (a) data remain normal while hack has happened. (b) SHAP Analysis}
    \label{fig_model}
\end{figure}

A second group of errors occurred when signals were indeed hacked, but traffic volumes remained low. SHAP plots revealed that in these mislabeled cases, none of the congestion characteristics (e.g., length of the jam, duration of the halt) stood out as abnormal enough to raise CNN's suspicion in Figure~\ref{fig_model}. This indicates that for stealth hacks in light traffic scenarios, our model struggles to detect abnormal signals when no clear congestion patterns arise. In these cases, the misclassification is likely due to the model's limitations in recognizing more subtle hacking indicators beyond standard congestion features. Model errors indicates that the algorithm must learn additional signals (e.g., irregular signal timing intervals) to catch attacks that do not manifest as obvious jams.

\section{Conclusion}

This study presents a machine learning-based anomaly detection system for identifying compromised traffic signals using only traffic flow data, addressing the challenge of limited access to network-layer information. Our analysis demonstrates that Longest Stop Duration and Total Jam Distance are key indicators of cyberattacks, as revealed through Explainable AI (XAI) techniques. Furthermore, misclassification analysis identifies two primary error sources: transitional data inconsistencies, where mislabeled recovery-phase traffic misleads the model, and model limitations, where stealth attacks in low-traffic conditions evade detection due to the lack of clear congestion patterns. These findings highlight the need for enhanced detection strategies that incorporate additional behavioral indicators to improve robustness against sophisticated cyber threats in traffic management systems.

% \section{Future Works}
% We aim to move forward with the following thrusts:
%     \begin{itemize}
%         \item Completing the network topology of traffic light controllers to attain a more realistic model. Currently, each traffic light is connected to is nearest cellular base tower, but in real scenarios, multiple cellular towers could serve the same lights. During the topological network phase, we would provide each traffic light controller the opportunity of connecting to the three nearest cellular towers.
%         \item The network virtualization in infrastructure simulation phase still relies highly on virtual machines, which creates a huge overhead. As a result, containerization of the project is a large priority to reduce computational cost. We aim to use Docker containers to orchestrate the network communication functionality of the framework.
%         \item During exploration of statistical learning approaches to identify cyber attacks in a supervise learning paradigm, it was found that statistical learning methods are insufficient to detect cyber attacks purely from traffic flow patterns. Deep Learning or Deep Unsupervised Learning are the next option to attempt to rectify this issue.
%     \end{itemize}

% \label{sectCC}

\section*{Acknowledgment}

This research was supported by the USDOT Tier-1 University Transportation Center (UTC) Transportation Cybersecurity Center for Advanced Research and Education (CYBER-CARE) (Grant No. 69A3552348332).

\bibliographystyle{IEEEtran}
% Tell the complier which style you want.
\bibliography{ref.bib}

\begin{thebibliography}{10}
\providecommand{\url}[1]{#1}
\csname url@samestyle\endcsname
\providecommand{\newblock}{\relax}
\providecommand{\bibinfo}[2]{#2}
\providecommand{\BIBentrySTDinterwordspacing}{\spaceskip=0pt\relax}
\providecommand{\BIBentryALTinterwordstretchfactor}{4}
\providecommand{\BIBentryALTinterwordspacing}{\spaceskip=\fontdimen2\font plus
\BIBentryALTinterwordstretchfactor\fontdimen3\font minus \fontdimen4\font\relax}
\providecommand{\BIBforeignlanguage}[2]{{%
\expandafter\ifx\csname l@#1\endcsname\relax
\typeout{** WARNING: IEEEtran.bst: No hyphenation pattern has been}%
\typeout{** loaded for the language `#1'. Using the pattern for}%
\typeout{** the default language instead.}%
\else
\language=\csname l@#1\endcsname
\fi
#2}}
\providecommand{\BIBdecl}{\relax}
\BIBdecl

\bibitem{vegesna2022investigations}
V.~V. Vegesna, ``Investigations on cybersecurity challenges and mitigation strategies in intelligent transport systems,'' \emph{Irish Interdisciplinary Journal of Science \& Research (IIJSR) Vol}, vol.~6, pp. 70--86, 2022.

\bibitem{feng2022cybersecurity}
Y.~Feng, S.~E. Huang, W.~Wong, Q.~A. Chen, Z.~M. Mao, and H.~X. Liu, ``On the cybersecurity of traffic signal control system with connected vehicles,'' \emph{IEEE Transactions on Intelligent Transportation Systems}, vol.~23, no.~9, pp. 16\,267--16\,279, 2022.

\bibitem{ozarpa2021cyber}
C.~{\"O}zarpa, {\.I}.~Avc{\i}, B.~F. K{\i}nac{\i}, S.~Arapo{\u{g}}lu, and S.~A. Kara, ``Cyber attacks on scada based traffic light control systems in the smart cities,'' \emph{The International Archives of the Photogrammetry, Remote Sensing and Spatial Information Sciences}, vol.~46, pp. 411--415, 2021.

\bibitem{pundir2022cyber}
A.~Pundir, S.~Singh, M.~Kumar, A.~Bafila, and G.~J. Saxena, ``Cyber-physical systems enabled transport networks in smart cities: Challenges and enabling technologies of the new mobility era,'' \emph{IEEE Access}, vol.~10, pp. 16\,350--16\,364, 2022.

\bibitem{feng2018vulnerability}
Y.~Feng, S.~Huang, Q.~A. Chen, H.~X. Liu, and Z.~M. Mao, ``Vulnerability of traffic control system under cyberattacks with falsified data,'' \emph{Transportation research record}, vol. 2672, no.~1, pp. 1--11, 2018.

\bibitem{zhou2022survey}
Y.~Zhou, D.~Liu, and H.~Song, ``A survey of machine learning algorithms and techniques for air mobility under emergency situations,'' in \emph{Proceedings of the 2022 IEEE International Conferences on Internet of Things (iThings), Green Computing \& Communications (GreenCom), Cyber, Physical \& Social Computing (CPSCom), Smart Data (SmartData), and Congress on Cybermatics (Cybermatics)}.\hskip 1em plus 0.5em minus 0.4em\relax IEEE, 2022, pp. 582--588.

\bibitem{ma2021smart}
C.~Ma, ``Smart city and cyber-security; technologies used, leading challenges and future recommendations,'' \emph{Energy Reports}, vol.~7, pp. 7999--8012, 2021.

\bibitem{gu2007denial}
Q.~Gu and P.~Liu, ``Denial of service attacks,'' \emph{Handbook of Computer Networks: Distributed Networks, Network Planning, Control, Management, and New Trends and Applications}, vol.~3, pp. 454--468, 2007.

\bibitem{chen2016analysis}
X.~Chen, Y.~Zhou, H.~Zhou, C.~Wan, Q.~Zhu, W.~Li, and S.~Hu, ``Analysis of production data manipulation attacks in petroleum cyber-physical systems,'' in \emph{2016 IEEE/ACM International Conference on Computer-Aided Design (ICCAD)}.\hskip 1em plus 0.5em minus 0.4em\relax IEEE, 2016, pp. 1--7.

\bibitem{syverson1994taxonomy}
P.~Syverson, ``A taxonomy of replay attacks [cryptographic protocols],'' in \emph{Proceedings The Computer Security Foundations Workshop VII}.\hskip 1em plus 0.5em minus 0.4em\relax IEEE, 1994, pp. 187--191.

\bibitem{usama2024command}
M.~Usama and M.~N. Aman, ``Command injection attacks in smart grids: A survey,'' \emph{IEEE Open Journal of Industry Applications}, 2024.

\bibitem{conti2016survey}
M.~Conti, N.~Dragoni, and V.~Lesyk, ``A survey of man in the middle attacks,'' \emph{IEEE communications surveys \& tutorials}, vol.~18, no.~3, pp. 2027--2051, 2016.

\bibitem{bhuyan2013network}
M.~H. Bhuyan, D.~K. Bhattacharyya, and J.~K. Kalita, ``Network anomaly detection: methods, systems and tools,'' \emph{Ieee communications surveys \& tutorials}, vol.~16, no.~1, pp. 303--336, 2013.

\bibitem{corea2024explainable}
P.~M. Corea, Y.~Liu, J.~Wang, S.~Niu, and H.~Song, ``Explainable ai for comparative analysis of intrusion detection models,'' \emph{arXiv preprint arXiv:2406.09684}, 2024.

\bibitem{liu2021zero}
Y.~Liu, J.~Wang, J.~Li, S.~Niu, L.~Wu, and H.~Song, ``Zero-bias deep-learning-enabled quickest abnormal event detection in iot,'' \emph{IEEE Internet of Things Journal}, vol.~9, no.~13, pp. 11\,385--11\,395, 2021.

\bibitem{liu2021class}
Y.~Liu, J.~Wang, J.~Li, S.~Niu, and H.~Song, ``Class-incremental learning for wireless device identification in iot,'' \emph{IEEE Internet of Things Journal}, vol.~8, no.~23, pp. 17\,227--17\,235, 2021.

\bibitem{pang2021deep}
G.~Pang, C.~Shen, L.~Cao, and A.~V.~D. Hengel, ``Deep learning for anomaly detection: A review,'' \emph{ACM computing surveys (CSUR)}, vol.~54, no.~2, pp. 1--38, 2021.

\bibitem{jyothi2024data}
V.~Jyothi, T.~Sreelatha, T.~Thiyagu, R.~Sowndharya, and N.~Arvinth, ``A data management system for smart cities leveraging artificial intelligence modeling techniques to enhance privacy and security,'' \emph{Journal of Internet Services and Information Security}, vol.~14, no.~1, pp. 37--51, 2024.

\bibitem{bawaneh2019anomaly}
M.~Bawaneh and V.~Simon, ``Anomaly detection in smart city traffic based on time series analysis,'' in \emph{2019 International Conference on Software, Telecommunications and Computer Networks (SoftCOM)}.\hskip 1em plus 0.5em minus 0.4em\relax IEEE, 2019, pp. 1--6.

\bibitem{zeiler2014visualizing}
\BIBentryALTinterwordspacing
M.~D. Zeiler and R.~Fergus, ``Visualizing and understanding convolutional networks,'' in \emph{European Conference on Computer Vision (ECCV)}.\hskip 1em plus 0.5em minus 0.4em\relax Springer, 2014, pp. 818--833. [Online]. Available: \url{https://arxiv.org/abs/1311.2901}
\BIBentrySTDinterwordspacing

\bibitem{ribeiro2016why}
\BIBentryALTinterwordspacing
M.~T. Ribeiro, S.~Singh, and C.~Guestrin, ``"why should i trust you?": Explaining the predictions of any classifier,'' in \emph{Proceedings of the 22nd ACM SIGKDD International Conference on Knowledge Discovery and Data Mining (KDD)}.\hskip 1em plus 0.5em minus 0.4em\relax ACM, 2016, pp. 1135--1144. [Online]. Available: \url{https://arxiv.org/abs/1602.04938}
\BIBentrySTDinterwordspacing

\end{thebibliography}
% Tell the compiler where to look for your references, IPCCC2022.bib here is the database that contains reference entries.

\end{document}